\pdfoutput=1

\documentclass[11pt]{article}

\usepackage[final]{acl}
\usepackage{geometry}
\usepackage{longtable}
\usepackage{multicol}
\usepackage{multirow}
\usepackage{times}
\usepackage{latexsym}
\usepackage{comment}
\usepackage[T1]{fontenc}

\usepackage[utf8]{inputenc}
\usepackage{hyperref}
\usepackage{microtype}

\usepackage{inconsolata}
\usepackage{avant}
\usepackage[T1]{fontenc}

\usepackage{graphicx}

\title{\textsc{InsightBuddy-AI}: Medication Extraction and Entity Linking \\ using Large Language Models and Ensemble Learning}

\author{Pablo Romero \\
  Manchester Metropolitan University \\
  Oxford Rd \\
  Greater Manchester, UK\\
  \texttt{pablo2004romero@gmail.com} \\\And
  Lifeng Han$^*$ \\
  LIACS \& LUMC\\ Leiden University, NL\\ 
  University of Manchester, UK\\
  \texttt{l.han@lumc.nl} \\ $^*$ \textit{corresponding author}\\[0.5em]
  \href{https://github.com/pabloRom2004/Insight-Buddy-AI-App}{{\fontfamily{pag}\selectfont\color[HTML]{4B9CD3}Software Download Link}}\\
  \And Goran Nenadic \\
  University of Manchester\\
  Oxford Rd \\
  Greater Manchester, UK\\
  \texttt{g.nenadic@manchester.ac.uk} \\ }

\begin{document}
\maketitle

\begin{abstract}
Medication Extraction and Mining play an important role in healthcare NLP research due to its practical applications in hospital settings, such as their mapping into standard clinical knowledge bases (SNOMED-CT, BNF, etc.).
In this work, we investigate state-of-the-art LLMs in text mining tasks on medications and their related attributes such as dosage, route, strength, and adverse effects. In addition, we explore different ensemble learning methods (\textsc{Stack-Ensemble} and \textsc{Voting-Ensemble}) to augment the model performances from individual LLMs. Our ensemble learning result demonstrated better performances than individually fine-tuned base models BERT, RoBERTa, RoBERTa-L, BioBERT, BioClinicalBERT, BioMedRoBERTa, ClinicalBERT, and PubMedBERT across general and specific domains. 
Finally, we build up an entity linking function to map extracted medical terminologies into the SNOMED-CT codes and the British National Formulary (BNF) codes, which are further mapped to the Dictionary of Medicines and Devices (dm+d), and ICD.
\end{abstract}

\section{Introduction}
Information Extraction on Medications and their related attributes plays an important role in natural language processing (\textbf{NLP}) applications in the \textbf{clinical} domain to support digital healthcare. 
Clinicians and healthcare professionals have been doing manual clinical \textbf{coding} for quite a long time to map clinical events such as diseases, drugs, and treatments into the existing terminology knowledge base, for instance, ICD and SNOMED. The procedure can be time-consuming yet without a guarantee of total correctness due to human-introduced errors. 
With the process of automated information extraction on \textbf{medications}, it will be further possible to automatically map the extracted terms into the current terminology database, i.e. the automated clinical coding. 
Due to the promising future of this procedure, different NLP models have been deployed in medication mining and clinical coding in recent years. However, they are often studied separately. In this work, 1) we investigate text mining of medications and their related attributes (dosage, route, strength, adverse effect, frequency, duration, form, and reason) together with \textit{automated clinical coding} into one pipeline. 
In addition, 2) we investigate the \textbf{ensemble} learning mechanisms (Stack and Voting) on a broad range of NLP models fine-tuned for named entity recognition (NER) tasks. These models include both general domain trained BERT, RoBERTa, RoBERTa-L, and domain-specific trained BioBERT, BioClinicalBERT, BioMedRoBERTa, ClinicalBERT, and PubMedBERT.
In this way, users do not have to worry about which models to choose for clinical NER. Instead, they can just place the newer models into the ensemble-learning framework to test their performances.

\section{Literature Review and Related Work}


Named Entity Recognition (\textbf{NER}) is a critical task for extracting key information from unstructured text, like medical letters. The complexity and context-dependency of medical language pose significant challenges for accurate entity extraction. Traditional approaches to NER, such as rule-based systems, have shown limited success in capturing the nuanced contextual information crucial for clinical NER \cite{nadeau2007survey}.
The advent of deep learning methods, particularly Long Short-Term Memory (LSTM) networks, marked a significant improvement in NER performance \cite{GRAVES2005602LSTM}, e.g. the ability to capture long-range dependencies in text. 
However, these models still struggled with rare entities and complex contextual relationships in \textbf{clinical notes}.
The introduction of BERT (Bidirectional Encoder Representations from Transformers) \cite{devlin-etal-2019-bert} 
revolutionised various NLP tasks, including NER. 
The model's pre-training on a large corpus using a masked language modelling objective builds rich token representations. The model can then be later fine-tuned by adding a classification layer at the end of the network to make decisions over each individual token embedding.

However, BERT's pre-training on general domain corpora (Wikipedia and books) limited its effectiveness on specialised medical texts. This limitation led to the development of \textbf{domain-specific} BERT variants. For example, BioBERT \cite{10.1093/bioinformatics/biobert}, pre-trained on large-scale biomedical corpora; ClinicalBERT \cite{wang2023optimized_clinicalBERT}, fine-tuned on EHR data from 3 million patients after pre-training on 1.2 billion words of diverse diseases, and other variants like Med-BERT \cite{rasmy2021medBERT} have demonstrated enhanced performance on medical NER tasks due to their specialised training on the medical domain \footnote{there have been other versions of Clinical BERTs such as \cite{clinicalbert2019huang} and \cite{alsentzer-etal-2019-publicly_bioclinicalBERT} that were trained on Medical Information Mart for Intensive Care III (mimiciii) data \cite{johnson2016mimic} respectively.}. 

Despite these improvements, \textbf{single-model approaches still struggle} with the inherent complexity and variability of clinical text, as the comparative studies reported in \cite{TransformerCRF_2023} across different models using BERT, ClinicalBERT, BioBERT, and scratch-learned Transformers. \textbf{Ensemble} methods have emerged as a promising direction to address these challenges, they have proven useful in other fields, such as computer vision \cite{Lee-etal-2018Ensemble}. By combining multiple models, ensembles can leverage the strengths of different models while mitigating their individual weaknesses. In the context of NER, ensembling has shown performance improvements, as shown by \cite{naderi2021ensemble}, where an ensemble is used on a health and life science corpora for a significant improvement in performance over single models. 
\newcite{naderi2021ensemble} conducted max voting for word-level biology, chemistry, and medicine data. 
However, 
on clinical/medical NER, they only focused on French using the DEFT benchmark dataset; while for the other two domains of biology and chemistry, they tested on English data. 
%
There are two commonly used ensemble methods, voting and stacked ensembles: 1)
\textbf{Maximum voting} in ensembles where each model contributes equally to the final decision as used in the paper \cite{naderi2021ensemble} have proved effective. This is where the most voted label is picked.
    2) Training a network on the outputs of the ensemble aims to capture more nuanced relationships. This is accomplished using a method called \textbf{stacking} introduced by \newcite{WOLPERT1992241stacked}. Stacking offers a more sophisticated approach by training a meta-model on the outputs of the base ensemble; the model is expected to learn more complex patterns from the ensemble outputs, leading to better predictions. This has proven effective in this paper \cite{app2022stacking-arabic} where they use a stacked ensemble with a support vector machine (SVM) for \textit{sentiment} analysis. Instead, we will use a simple feed-forward network from the outputs of the ensemble to the final labels for our tasks. more examples on stacked ensemble can be found at \cite{MOHAMMED20228825ensemble,gunecs2017stacked}.

While ensemble methods have shown promise in other NER tasks, their effectiveness for clinical NER, particularly on challenging datasets like n2c2 2018 \cite{henry20202018_n2c2_task2}, remains unexplored. 
%
This work \textbf{aims to address this gap} by investigating \textit{whether stacked and voting ensembles can make an impact on NER tasks on clinical notes}, potentially overcoming the limitations of current single-model approaches.

\begin{figure*}[th]
\centering
\includegraphics[width=0.99\textwidth]{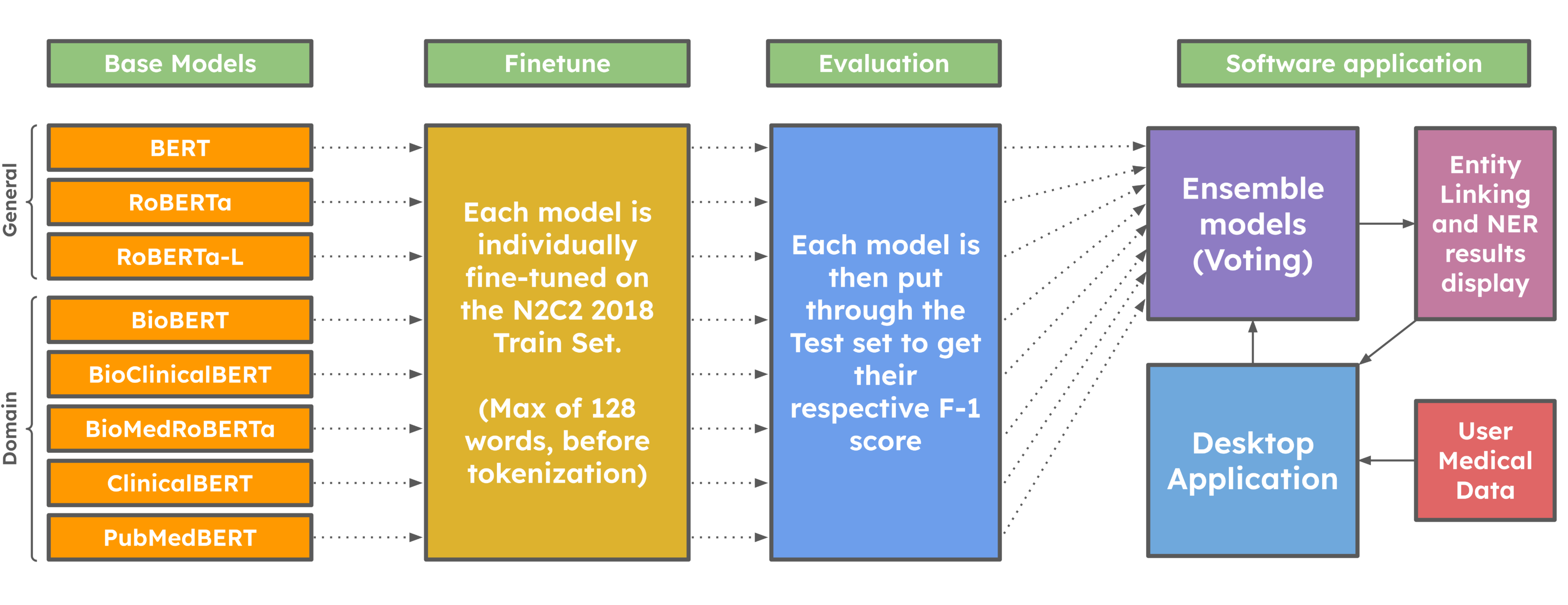}
    \caption{\textsc{InsightBuddy} Framework Pipeline: individual NER model fine-tuning, ensemble, and entity linking. Two kinds of base models include the general domain and the biomedical domain with their Huggingface repositories in Table \ref{tab:EnsembleNER-model-list}. Pre-preprocessing data: cut the sequence with the first full stop ``.'' after the 100th token, otherwise, cut the sequence up to 128 tokens. Fine-tuning: using the same parameter sets for all eight models. Ensemble: different strategies will be displayed in Fig \ref{fig:ensemble-NER-only-diag}. Entity Linking: links to clinical KB including BNF and SNOMED.}
    \label{fig:ensemble-NER-pipe}
\end{figure*}

\section{Methodologies}
The Overall framework of \textsc{InsightBuddy} is shown in Figure \ref{fig:ensemble-NER-pipe}, which displays the base models we included from the general domain 1) BERT \cite{devlin-etal-2019-bert}, RoBERTa \cite{DBLP:journals/corr/abs-1907-11692Roberta}, and RoBERTa-Large, and 2)  biomedical/clinical domains BioBERT \cite{10.1093/bioinformatics/biobert}, BioClinicalBERT \cite{alsentzer-etal-2019-publicly_bioclinicalBERT}, BioMedRoBERTa \cite{biomedRoBERTa2020}, ClinicalBERT \cite{wang2023optimized_clinicalBERT}, and PubMedBERT \cite{pubmedbert2020}. 
The fine-tuning of eight models uses the same set of parameters (Section \ref{sec:appendix_param} for parameter selections) and the n2c2-2018 shared task training data with data pre-processing. 
The initial evaluation phase using n2c2-2018 testing set gives an overall idea of each model's performance. This is followed by ensemble learning on all the models' outputs. 
With the output from NER models, we add an \textbf{entity linking} function to map the extracted medical entities into the standard clinical terminology knowledge base (KB), using \textbf{SNOMED-CT and BNF} as our initial KB, which is further mapped to ICD and dm+d. 

For data pre-processing, we chunk the sequence into a maximum of 128 tokens. If there is a full stop ``.'' between the 100th  and 128th tokens, it will be cut at the full stop.
Regarding ensemble-learning strategy, we draw a InsightBuddy Ensemble figure (Figure \ref{fig:ensemble-NER-only-diag}) to explain in detail.
Firstly the initial output of eight individual fine-tuned NER models is tokenised, i.e. at the \textbf{sub-word} level, due to the model learning strategy, e.g. ``Para \#\#ce \#\#tam \#\#ol'' instead of ``Paracetamol''. What we need to do at the first step is to \textbf{group} the sub-word tokens into words for both practical application and voting purposes. However, each sub-word is labeled with predefined labels and these labels often do not agree with each other within the same words. We designed \textbf{three group solutions}, i.e. first-token voting/selection, max-token voting, and average voting. The \textit{first-token voting} is to assign a word the same label as its first sub-word piece. For example, using this strategy, the word ``Paracetamol'' will be labeled as ``B-Drug'' if its first sub-word ``Para'' is labeled as ``B-Drug'' regardless of other labels from the subsequent sub-words. 
The \textit{max-token voting} will assign a word the label that has the highest sub-word logit, this indicates that the model is more confident in that prediction, the higher the logit is. 
The \textit{average voting} solution calculates the average logits across all sub-words predictions and then samples from this to get the label for the entire word.

Regarding \textbf{word-level ensemble} learning, we investigate the classical \textbf{voting} strategy with modifications (two solutions).
For the first solution ``>=4 or O'', if there are more than half of the models agree on one label, we pick this label, i.e. >=4 such same labels. Otherwise, we assign the default ``O'' label to indicate it as context words, due to the models' disagreement.
For the second solution, we use max-voting, i.e. the most agreed label regardless of how many models they are, e.g. 2, 3, 4, or more. In this case, if there are ties, e.g. (3, 3, 2) two labels are voted both three times from six models, we need to decide on the tied labels. There are two solutions for the selection, 1) alphabetical, and 2) fully randomised.

We also draw the \textsc{\textbf{Stacked}-Ensemble} in Figure \ref{fig:stack-ensemble-pipe-train} and \ref{fig:stack-ensemble-pipe}, where the model training and one-hot encoded model predictions are illustrated. 
In the training phase, we cut the real data into 80\% and 20\% for the training and testing of the model. Model exports are conducted only if at least 2 models are predicting a label that is not ``O''; otherwise ``O'' is the default option and the output is ignored and not included in the stacked training data.
For training data collection, output logits for each model are converted into a one-hot encoded vector, concatenated and saved along with the real label for each token. There are 8 one-hot encoded vectors from 8 individual models and 1 label. So the model during training will see the value ``1'' eight times from the eight models, and the value ``0'' for the rest of the vector values. Overall, there are  8 vectors with each length of 19 digits. 
So there will be 8 (number of models) $\times$ 19 (number of labels) - 8 (eight 1s as there are 8 one hot encoded vectors so they have a single 1 each) = 144 ``0'' values
for every training example. 
We use \textit{one-hot encoding} instead of the output logits themselves to avoid the model \textit{overfitting} because the model makes more confident predictions when running on the training set. 
As this is the data that it was originally trained on, it is very confident with it's predictions. 
We can mitigate this by only feeding the one-hot encoded vectors to the stacked network. 


\section{Experimental Evaluations}

We use the n2c2-2018 shared task data on NER of adverse drug events and related medical attributes \cite{henry20202018_n2c2_task2}. The data is labeled with the following list of labels: ADE, Dosage, Drug, Duration, Form, Frequency, Reason, Route, and Strength in BIO format. So, overall, we have 19 labels, 2 (B/I) x 9 + 1 (O). The original training and testing sets are 303 and 202 letters respectively. We divided the original training set into two parts (9:1 ratio) for our model selection purposes: our new training and validation set, following the data split from recent work by \newcite{TransformerCRF_2023}.

We report Precision, Recall, and F1 score in two categories ``macro'' and ``weighted'', in addition to Accuracy. The ``\textbf{macro}'' category treats each label class the same weight regardless of their occurrence rates, while the ``\textbf{weighted}'' category'' assigns each label class with a weight according to their occurrence in the data. 
We first report the individual model fine-tuning scores and compare them with related work (subword level); then we report the ensemble model evaluation with different ensemble solutions (word level).

\subsection{Individual Models: sub-word level}

The performance of individual models after fine-tuning is reported in Table \ref{tab:individual-sub-word-level-models} 
where it says that RoBERTa-L performs the best in the macro Precision (0.8489), Recall (0.8606) and F1 (0.8538) score across general domain models, also winning domain-specific models.
BioiMedRoBERTa wins the domain-specific category models producing macro Precision, Recall, and F1 scores (0.8482 0.8477 0.8468).
In comparison to the NER work from \newcite{TransformerCRF_2023}, who's macro avg scores are: 0.842, 0.834, 0.837 from ClinicalBERT-Apt, our fine-tuned ClinicalBERT has similar performances (0.848, 0.825, 0.834), which shows our fine-tuning was successful. 
However, our best domain-specific model BioMedRoBERTa produces \textbf{higher} scores: macro
P/R/F1 (0.8482 0.\textbf{8477} 0.\textbf{8468}) and weighted P/R/F1 (0.9782 0.9775 0.9776) and Accuracy 0.9775 as in Figure 6.
Furthermore, the fine-tuned RoBERTa-L even achieved higher scores of (\textbf{0.8489 0.8606 0.8538}) for macro P/R/F1 and Acc 0.9782 in Figure 13. 
Both fine-tuned BioMedRoBERTa and RoBERTa-Large also \textit{win the best models} reported by \newcite{TransformerCRF_2023} which is their  ClinicalBERT-CRF model, macro avg (0.85, 0.829, 0.837), Acc 0.976.
Afterwards, in this paper, we emphasis on \textbf{word level} instead of sub-word, which was focused on by \newcite{TransformerCRF_2023}.

\subsection{Ensemble: word-level grouping (logits)}
We tried \textbf{first} logit voting, \textbf{max} voting, and \textbf{average} voting to group sub-words into words with corresponding labels. Their results are shown in Table \ref{tab:Ensemble-votings-vs-stacked-word-level}, in the upper group. 
First logit voting produced a higher Recall 0.8260 while Max logit voting produced a higher Precision 0.8261 resulting in higher F1 0.8232, i.e. \textit{Max} logit > \textit{First} logit > \textit{Average} logit with macro F1 (0.8232, 0.8229, 0.8227). 
However, overall, their performance scores are very close, so we chose the first-logit voting output for the afterward word-level ensemble due to computational convenience. 

\subsection{Ensemble: Voting vs Stacked (one-hot)}

Regarding Stacked Ensemble using one-hot encoded vectors, as shown in the middle group in Table \ref{tab:Ensemble-votings-vs-stacked-word-level}, it actually produced higher Precision score 0.8351 in comparisons to the highest Precision 0.8261 from Voting Ensembles. However, the Recall score on macro avg is 2 point lower than the voting ensemble, 0.8065 vs 0.8260, which means that the Stacked Ensemble \textit{reduced the false positive errors} but also increased the false negative error prediction. This implies that it has stricter constraint on positive predictions.

\subsection{Ensemble Models: BIO-span vs non-strict word-level}
So far, we have been reporting the evaluation scores on the BIO-strict label categorization, i.e. we distinguish between the label's beginning or the inner part of the label. For instance, a B-Drug will be different from an I-Drug and it will be marked as wrong if they are different from the reference. 
However, we think in practice, there are situations when users do not need the BIO, especially B and I difference. In Table \ref{tab:Ensemble-votings-vs-stacked-word-level}, we can see that, without considering the label difference of B and I, only focusing on the 9 label categories, word level ensemble model produced much higher Macro avg evaluations cores on Precision (0.8844) and Recall (0.8830) leading to higher F1 (0.8821), in comparison to BI-distinguished Macro F1 0.8232 (voting-max-logit) and F1 0.8156 (stacked-first-logit).

\subsection{Word-level: voting ensembles vs individual fine-tuned}

As in Table \ref{tab:individual-vs-ensemble-max-voting-word-level},
BioMedRoBERTa individual word level max logit grouping scores macro avg  P/R/F1   (0.8065    0.8224    0.8122    563329) vs max logit ensemble voting P/R/F1 (0.8261    0.8259    0.8232), we can see that ensemble boosted P (0.8261-0.8065)/0.8065= 2.43\%, and F1 (0.8232-0.8122)/0.8122= 1.35\% which says the ensemble voting is successful. By increasing the Precision score, the \textit{ensembles reduce the \textbf{false positive} labels} in the system output, while keeping the Recall at the same level, i.e. the true positive labels.

\begin{table}[t]
    \centering
\begin{tabular}{|l|c|c|c|}
\hline
\multicolumn{4}{|c|}{\textbf{Voting Average Ensemble word level (BIO)}} \\
\hline
Metric         & P      & R      & F1     \\
\hline
accuracy       & \multicolumn{3}{c|}{0.9796} \\ \hline 
macro avg      & 0.8253 & 0.8256 & 0.8227 \\
weighted avg   & 0.9807 & 0.9796 & 0.9798 \\
\hline
\multicolumn{4}{|c|}{\textbf{Voting First logit Ensemble word level (BIO)}} \\
\hline
Metric         & P      & R      & F1     \\
\hline
accuracy       & \multicolumn{3}{c|}{0.9796} \\ \hline 
macro avg      & 0.8255 & \textbf{0.8260} & 0.8229 \\
weighted avg   & 0.9807 & 0.9796 & 0.9798 \\
\hline
\multicolumn{4}{|c|}{\textbf{Voting Max logit Ensemble word level (BIO)}} \\
\hline
Metric         & P      & R      & F1     \\
\hline
accuracy       & \multicolumn{3}{c|}{0.9796} \\ \hline 
macro avg      & 0.8261 & 0.8259 & \textbf{0.8232} \\
weighted avg   & 0.9807 & 0.9796 & 0.9798 \\
\hline\hline 
\multicolumn{4}{|c|}{\textbf{Stacked Ensemble first logit word level (BIO)}} \\
\hline
Metric         & P      & R      & F1     \\
\hline
accuracy       & \multicolumn{3}{c|}{0.9796} \\ \hline 
macro avg      & \textbf{0.8351} & 0.8065 & 0.8156 \\
weighted avg   & 0.9800 & 0.9796 & 0.9794 \\
\hline\hline 
\multicolumn{4}{|c|}{\textbf{Non-BIO-only-word ensemble}} \\
\hline
Metric         & P      & R      & F1     \\
\hline
accuracy       & \multicolumn{3}{c|}{0.9839} \\ \hline 
macro avg      & 0.8844 & 0.8830 & 0.8821 \\
weighted avg   & 0.9840 & 0.9839 & 0.9838 \\
\hline
\end{tabular}  \caption{word-level grouping ensemble voting, max > first > average logit voting though they are very close scores. The \textbf{stacked} ensemble has the highest \textbf{Precision} scores, but the lowest Recall scores, which lead to lower F1. In the bottom cluster, it is the word-level evaluation without distinguishing B/I labels,  evaluation on n2c2 2018 test data.}
    \label{tab:Ensemble-votings-vs-stacked-word-level}
\end{table}

\section{Entity Linking: BNF and SNOMED}
To map the identified named entities into the clinical knowledge base. We use the existing code mapping sheet from the British National Formulary (BNF) web between SNOMED-CT, BNF, dm+d, and ICD \footnote{\url{https://www.nhsbsa.nhs.uk/prescription-data/understanding-our-data/bnf-snomed-mapping}}.
We preprocessed the SNOMED code from 377,834 to 10,804 to filter repeated examples between the mapping of SNOMED and BNF. We looked for non-drug words present in the text, then we filtered the drugs further by seeing if words like ['system', 'ostomy', 'bag', 'filter', 'piece', 'closure'] were present in the text, and if so, it was discarded.

For SNOMED CT mapping, we applied a fuzzy search to the cleaned mapping list with drug names. Then the SNOMED CT code will be added to the searching function on the SNOMED CT web, whenever there is a match.
%
For BNF mapping, the linking function uses keyword search to retrieve the BNF website with corresponding drugs, due to its different searching features in comparison to the SNOMED-CT web page.
Potential users can select whichever is suitable to their preferences between the two clinical knowledge bases (KBs), Figure \ref{fig:linking-diag}. 

\section{Discussion and Conclusion}

In this paper, we investigated Stacked Ensemble and Voting Ensemble on medical named entity recognition tasks using eight pretrained LMs from both general and biomed/clinical domains.
Our experiments show that our fine-tuned best individual models outperformed the state-of-the-art on standard shared task data n2c2-2018. The two ensemble strategies using output logits and one-hot encoding further improved the model performances.
We also offer desktop applications and user interfaces for individual fine-tuned models where we added the entity linking/normalisation function to BNF and SNOMED CT clinical knowledge base. We call the package \textsc{InsightBuddy-AI}, which will be released upon acceptance. 

\section*{Limitations}
Ensemble methods, especially with large models, can be computationally intensive. 
We indeed encountered limitations regarding computational resources during training and inference. 
Future work includes addressing the computational overhead of ensembles, exploring other ensemble techniques, and applying the methods to different datasets.

\section*{Ethics}
To use the n2c2 shared task data,  the authors have carried out CITI training (\url{https://physionet.org/settings/credentialing/}) and gained the access to the data with user agreement.

\bibliography{custom}

\appendix

\section{InsightBuddy-AI Desktop Application}

\begin{figure*}[t]
\centering
\includegraphics[width=0.99\textwidth]{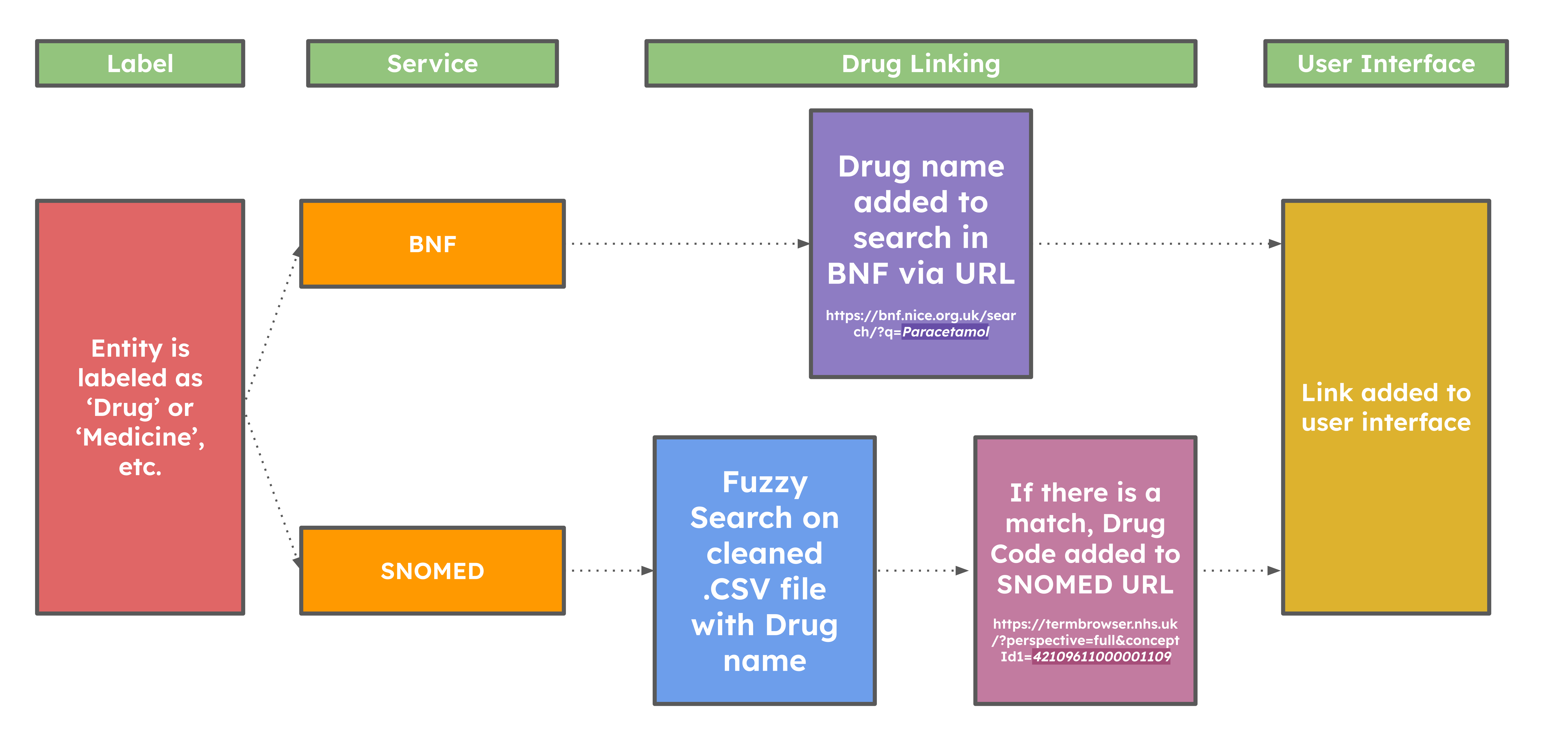}
    \caption{\textsc{EntityLinking}:  function illustration for mapping to both BNF and SNOMED-CT}
    \label{fig:linking-diag}
\end{figure*}

\begin{figure*}[]
\centering
\includegraphics[width=0.95\textwidth]{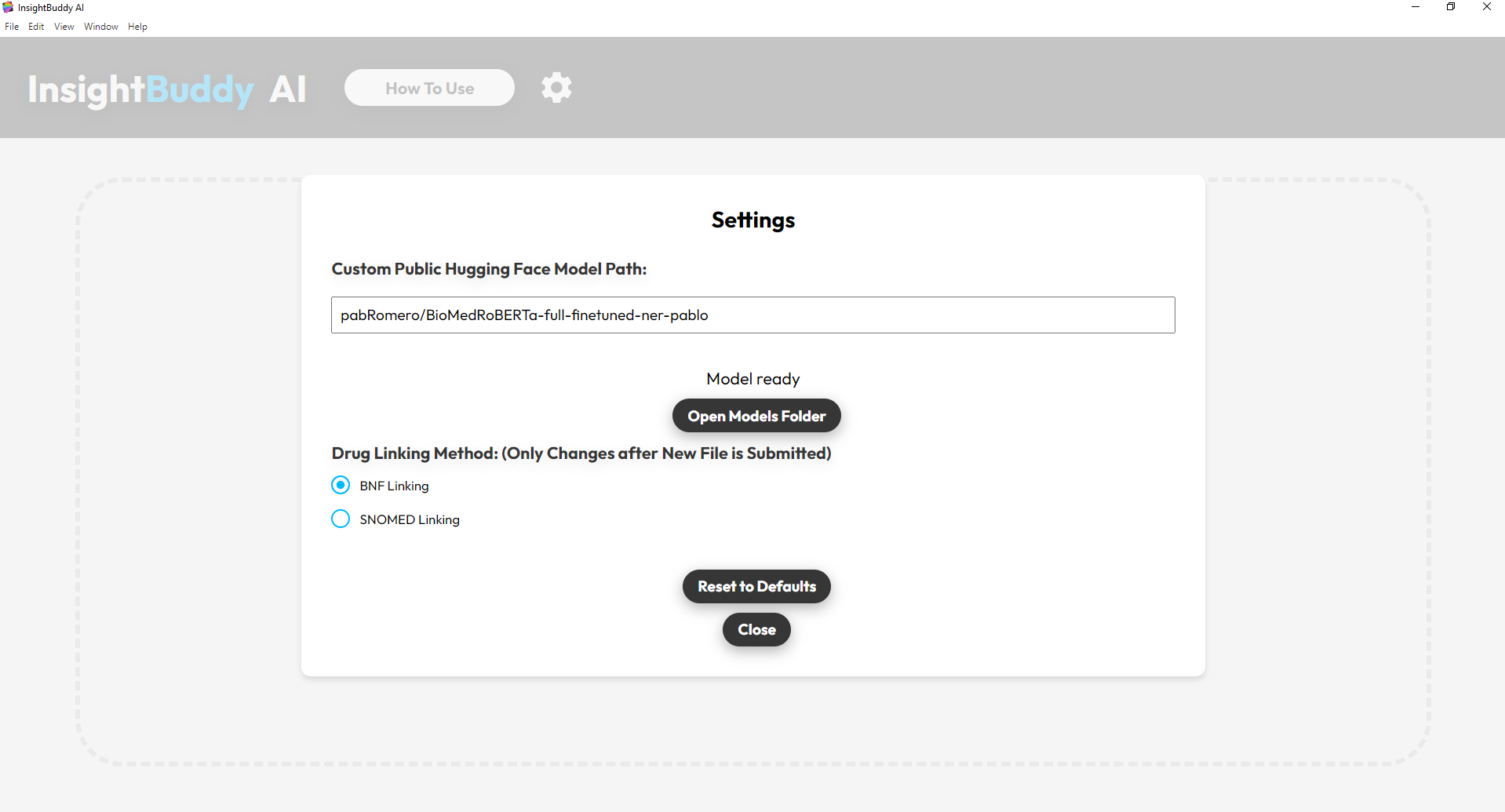}
    \caption{Choice of BNF and SNOMED-CT Linking}
    \label{fig:screenshot-any-BNF-SNOMED}
\end{figure*}

\begin{figure*}[]
\centering
\includegraphics[width=0.95\textwidth]{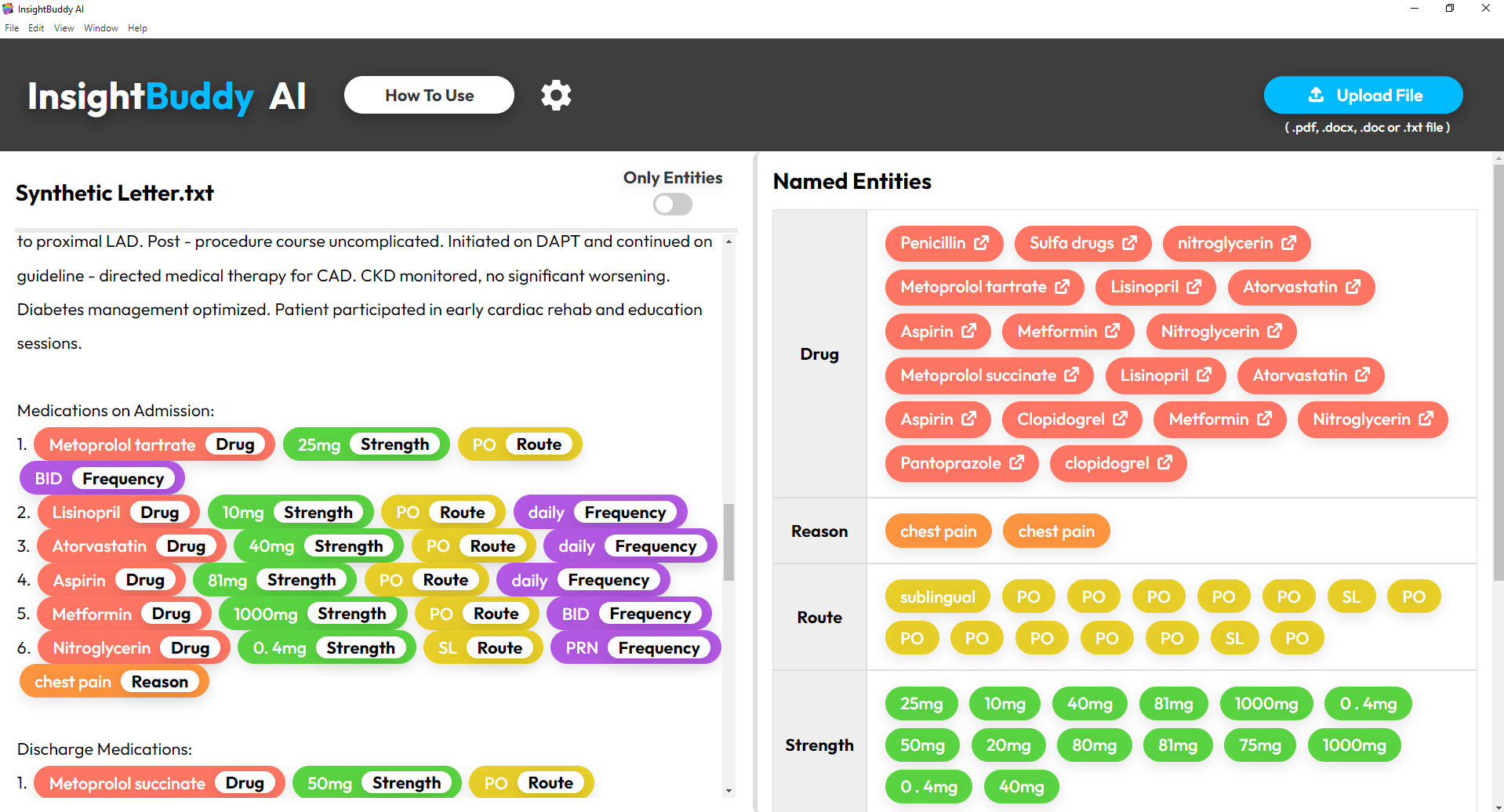}
    \caption{Demonstration of Clinical Events Outputs using A Synthetic Letter. }
    \label{fig:screenshot-syn-clinical-demo}
\end{figure*}

\begin{figure*}[]
\centering
\includegraphics[width=0.95\textwidth]{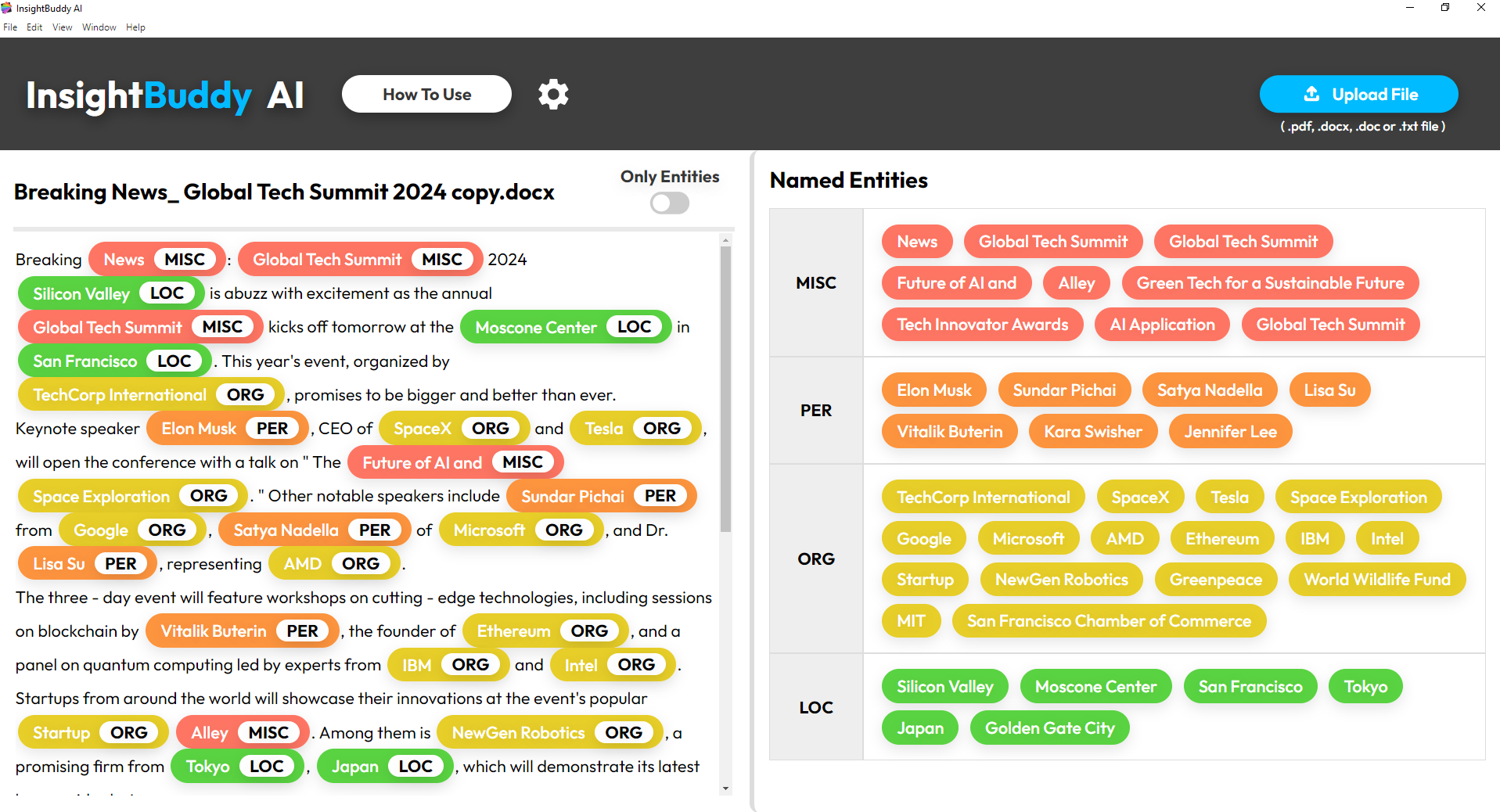}
    \caption{Loading Any Huggingface NER model: example outcome with typical (PER, LOC, ORG, MISC) label set}
    \label{fig:screenshot-any-huggingfaceNER}
\end{figure*}

\begin{figure*}[]
\centering
\includegraphics[width=0.95\textwidth]{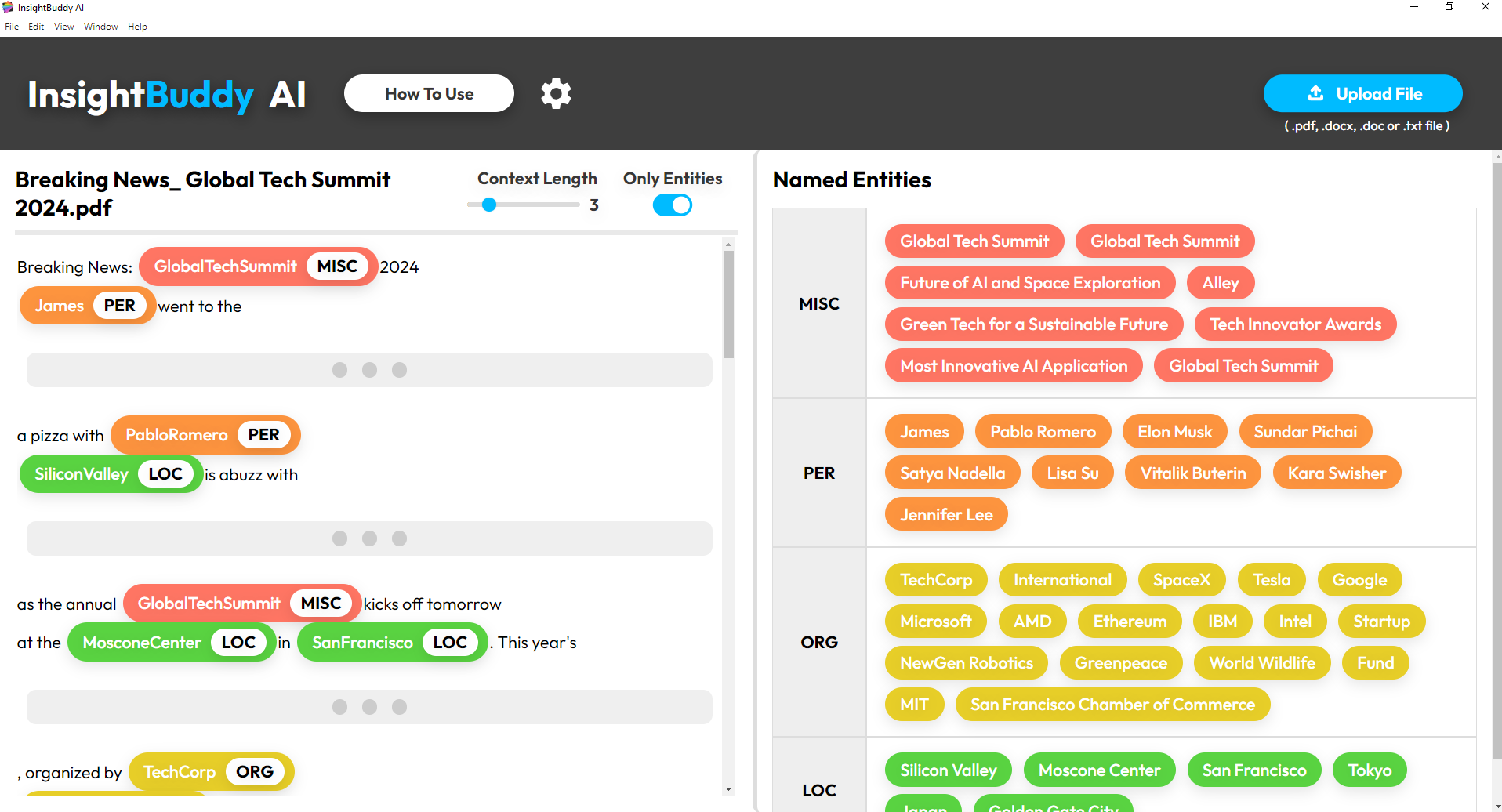}
    \caption{Context-awareness Feature using Window Parameter around the Entity}
    \label{fig:screenshot-feature-window}
\end{figure*}

We illustrate the Desktop Applications of InsightBuddy-AI in Figure \ref{fig:screenshot-syn-clinical-demo} and \ref{fig:screenshot-any-huggingfaceNER}, for demonstration of clinical event recognition using a synthetic letter via 1) loading our pre-trained model and common NER categories via 2) loading a Huggingface NER model.
There is also a \textbf{sliding window feature} called ``context length'' to allow flexible length of context around the entities visible to users, as in Figure \ref{fig:screenshot-feature-window}.
For \textbf{Clinical Coding} (entity linking) options, the desktop application can currently directly link the extracted entities to BNF and SNOMED-CT, as in Figure \ref{fig:screenshot-any-BNF-SNOMED} from the screenshots.
The \textsc{InsightBuddy-AI} software supports both Mac and Windows systems.

\section{Diagrams and Scoring Tables}
\label{sec:append_diagrams-scores}

\subsection{Word-level vs Sub-word Level scores}
From word-level ensemble result in Figure \ref{fig:max-Logit_Ensemble_Voting_word_bios}, it says that the ensembled model can achieve word-level evaluation scores 0.826, 0.826, and 0.823 for macro P/R/F1, which is close to sub-word level best model 0.847 F1. 
We can see that at word-level evaluation, there are 563,329 support tokens in Figure \ref{fig:max-Logit_Ensemble_Voting_word_bios}, vs sub-word level 756,014 tokens in Figure \ref{fig:BioMedRoBERTa_token_bios}. 

Word-level ensemble voting, max-logit voting > first-logit > average-logit, as shown in Table \ref{tab:Ensemble-votings-vs-stacked-word-level}, with Macro F1 scores (0.8232, 0.8229, 0.8227) respectively, which are very close though. They have the same weighted average F1 and Accuracy scores (0.9798, 0.9796) respectively.

\subsection{Ensemble: Stacked using output logits (non one-hot)}
When we used the `output logits' instead of `one-hot encoding' for stacked ensemble, as we discussioned in the methodology section, it will lead to overfitting issues. We use the Max logit stacked ensemble as an example, in figure \ref{fig:max_Stacked_Ensemble_word_bios}, which shows that the Stacked Ensemble using output logits produced much lower evaluation scores macro avg (0.6863    0.7339    0.6592) than the voting mechanism macro avg (0.8261    0.8259    0.8232) for (P, R, F1). The corresponding confusion matrix from the stacked ensemble using the max logit is shown in Figure \ref{fig:max_Stacked_Ensemble_word_confusion_matrix} with more errors spread in the image, the coloured numbers outside the diagonal line.

\begin{figure*}[]
\centering
\includegraphics[width=0.99\textwidth]{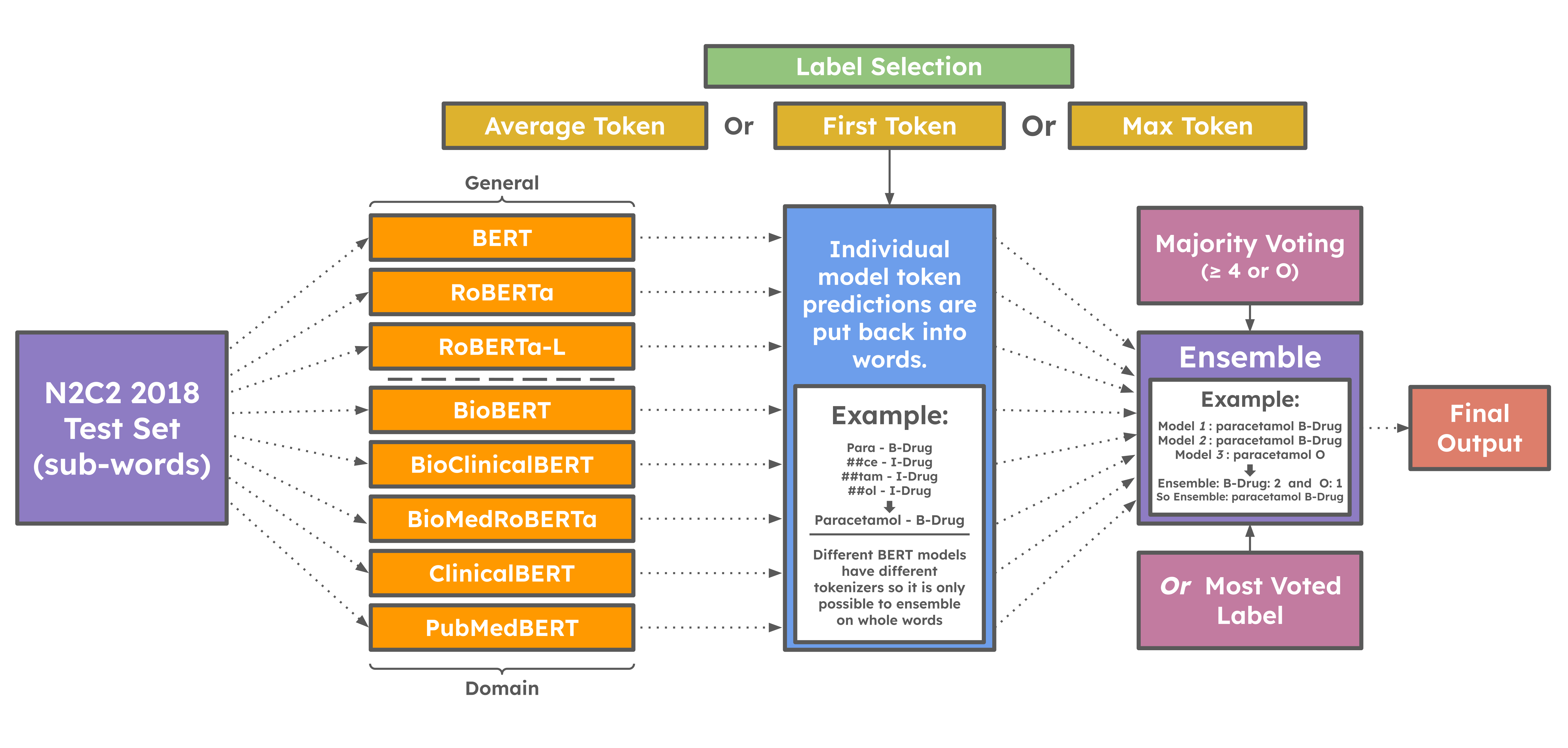}
   \caption{\textsc{InsightBuddy} Voted Ensemble Pipeline: individual NER model fine-tuning outputs are at token/sub-word level. "Logits are the outputs of a neural network before the activation function is applied" 
   first, we do the grouping of sub-words into words using three strategies: first token label, max token voting, or average voting (from our results, the first-token-lable selection gives higher Recall, while other two voting give higher precision, but they all end with the same F1 score, ref Table \ref{tab:Ensemble-votings-vs-stacked-word-level} ). || We take the best output from the first token label selection as the solution. For word-level ensemble on eight models, we have two solutions for voting, 1) either majority voting with >= 4 labels as the same then we pick it, otherwise choose default ``O'', or 2) max voting with the most popular label whatever it is; for max voting, if there is a tie, e.g. (3,3,2), we tested both alphabetical pick-up, or random pick-up of tied labels. Our results show that ``>=4, or O'' performs similarly to ``max + alphabetical'', while ``max + random'' slightly performs lower.}
    \label{fig:ensemble-NER-only-diag}
\end{figure*}

\begin{figure*}[t]
\centering
\includegraphics[width=0.99\textwidth]{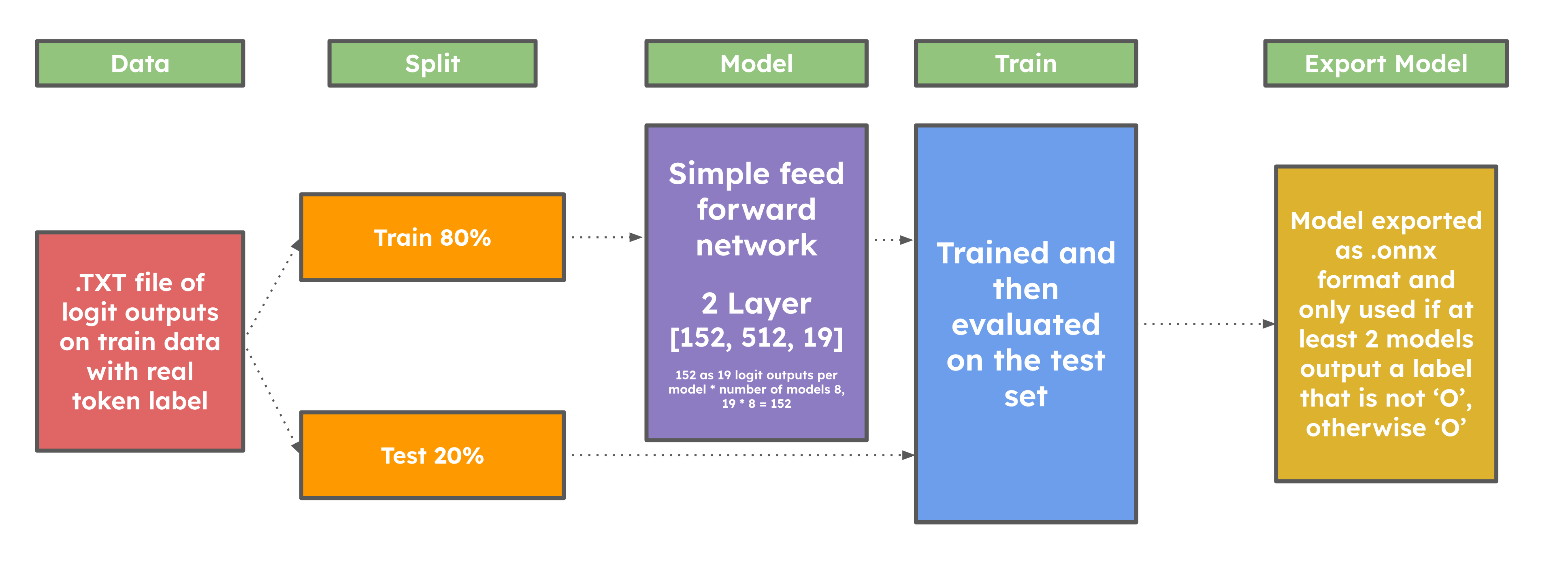}
    \caption{\textsc{StackedEnsemble}: training strategy.  }
    \label{fig:stack-ensemble-pipe-train}
\end{figure*}

\begin{figure*}[t]
\centering
\includegraphics[width=0.99\textwidth]{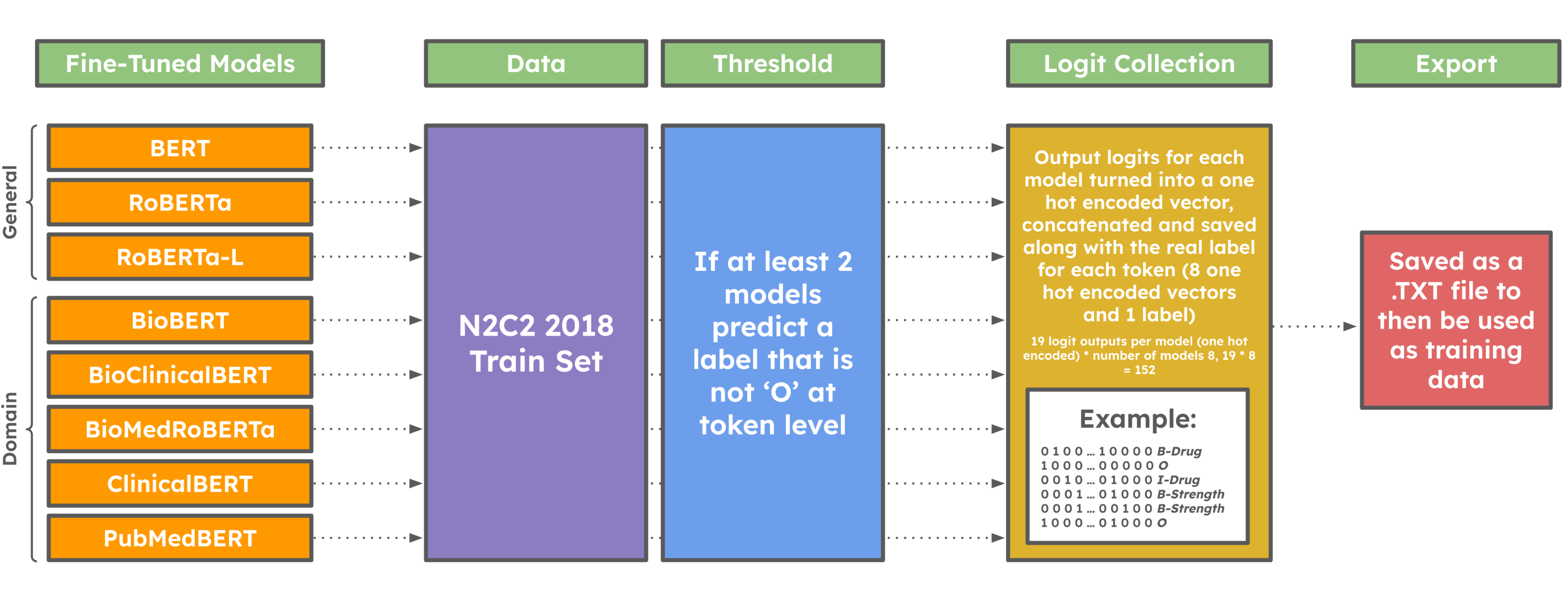}
    \caption{\textsc{StackedEnsemble}: one-hot encoding data. Instead of feeding the first logit, which produced a lower score due to overfitting, we now feed the one-hot encoding, which produced the highest Precision.  }
    \label{fig:stack-ensemble-pipe}
\end{figure*}

\begin{figure*}[]
\centering
\includegraphics[width=0.99\textwidth]{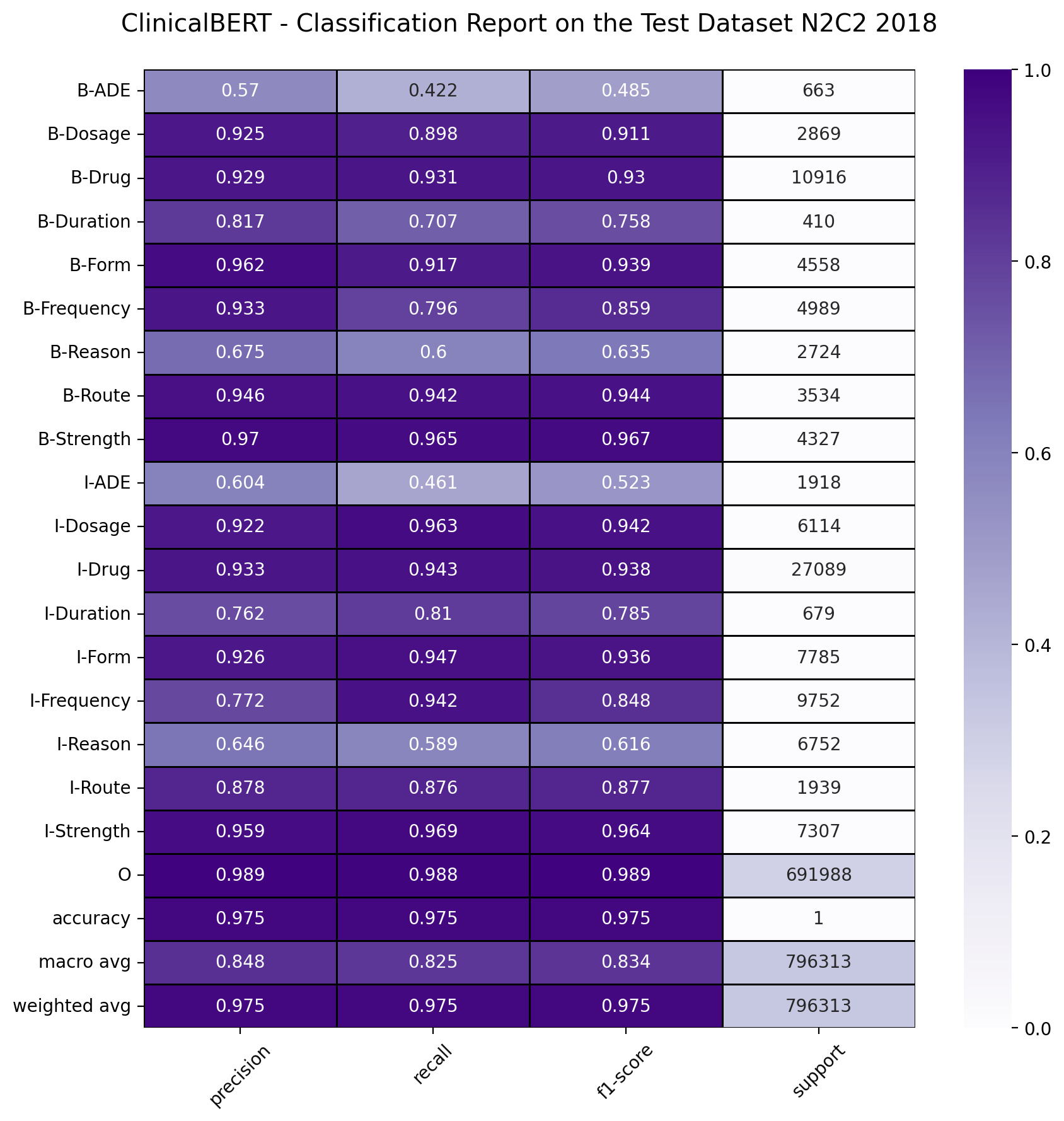}
    \caption{ClinicalBERT Eval at Sub-word Level. This score is similar to \cite{TransformerCRF_2023} paper on ClinicalBERT-Apt whose macro: .842, .834, .837; and weighted: .974, .974, .974) which says our fine-tuning is successful. \textbf{However, our best domain-specific model BioMedRoBERTa produces higher score: macro P/R/F (0.8482    0.8477    0.8468) and weighted P/R/F (0.9782    0.9775    0.9776) and Accuracy 0.9775 as in Figure \ref{fig:BioMedRoBERTa_token_bios}. Furthermore, the fine-tuned RoBERTa-L even achieved higher scores of (0.8489 0.8606 0.8538) for P/R/F1 and Acc 0.9782 in Table \ref{tab:individual-vs-ensemble-max-voting-word-level}}. Afterwards, in this paper, we emphasis on word level instead of sub-word, which was focused by \cite{TransformerCRF_2023}.}
    \label{fig:ClinicalBERT_token_bios}
\end{figure*}

\begin{figure*}[]
\centering
\includegraphics[width=0.99\textwidth]{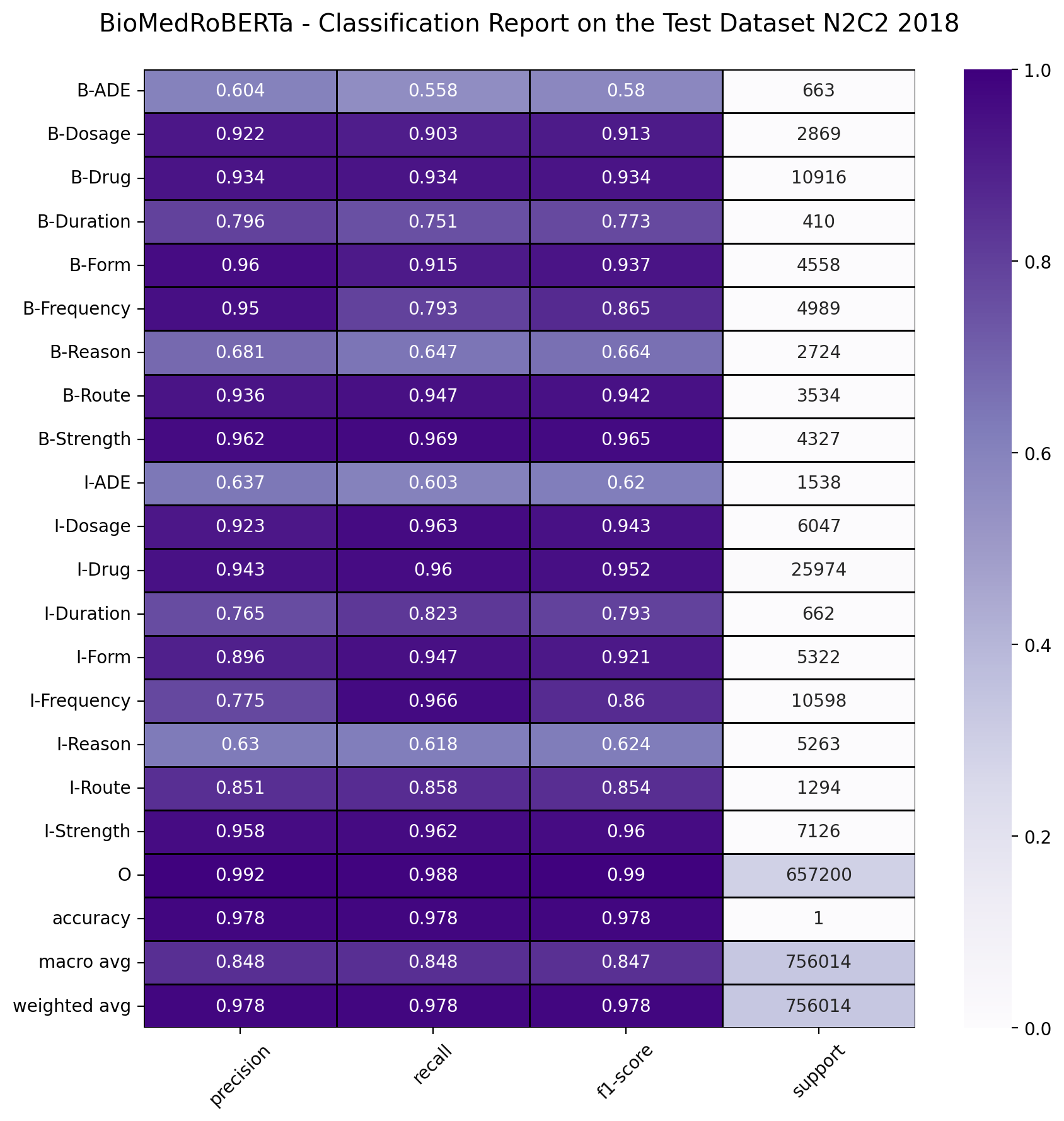}
    \caption{BioMedRoBERTa Eval at Sub-word Level on n2c2 2018 test data.}
    \label{fig:BioMedRoBERTa_token_bios}
\end{figure*}

\begin{figure*}[]
\centering
\includegraphics[width=0.99\textwidth]{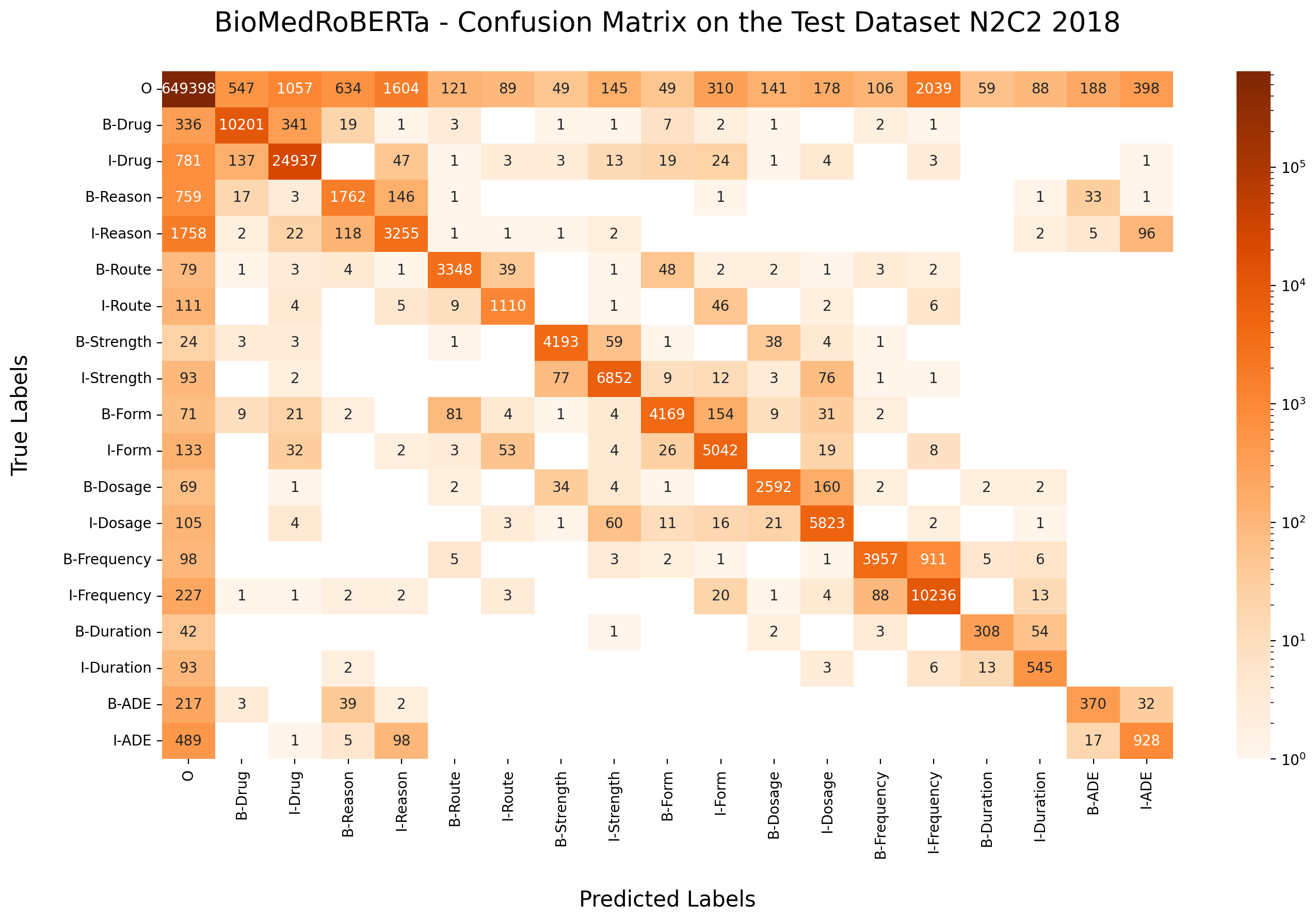}
    \caption{BioMedRoBERTa Eval Confusion Matrix at Sub-word Level on n2c2 2018 test data.}
    \label{fig:BioMedRoBERTa_token_confusion_matrix}
\end{figure*}

\begin{figure*}[]
\centering
\includegraphics[width=0.99\textwidth]{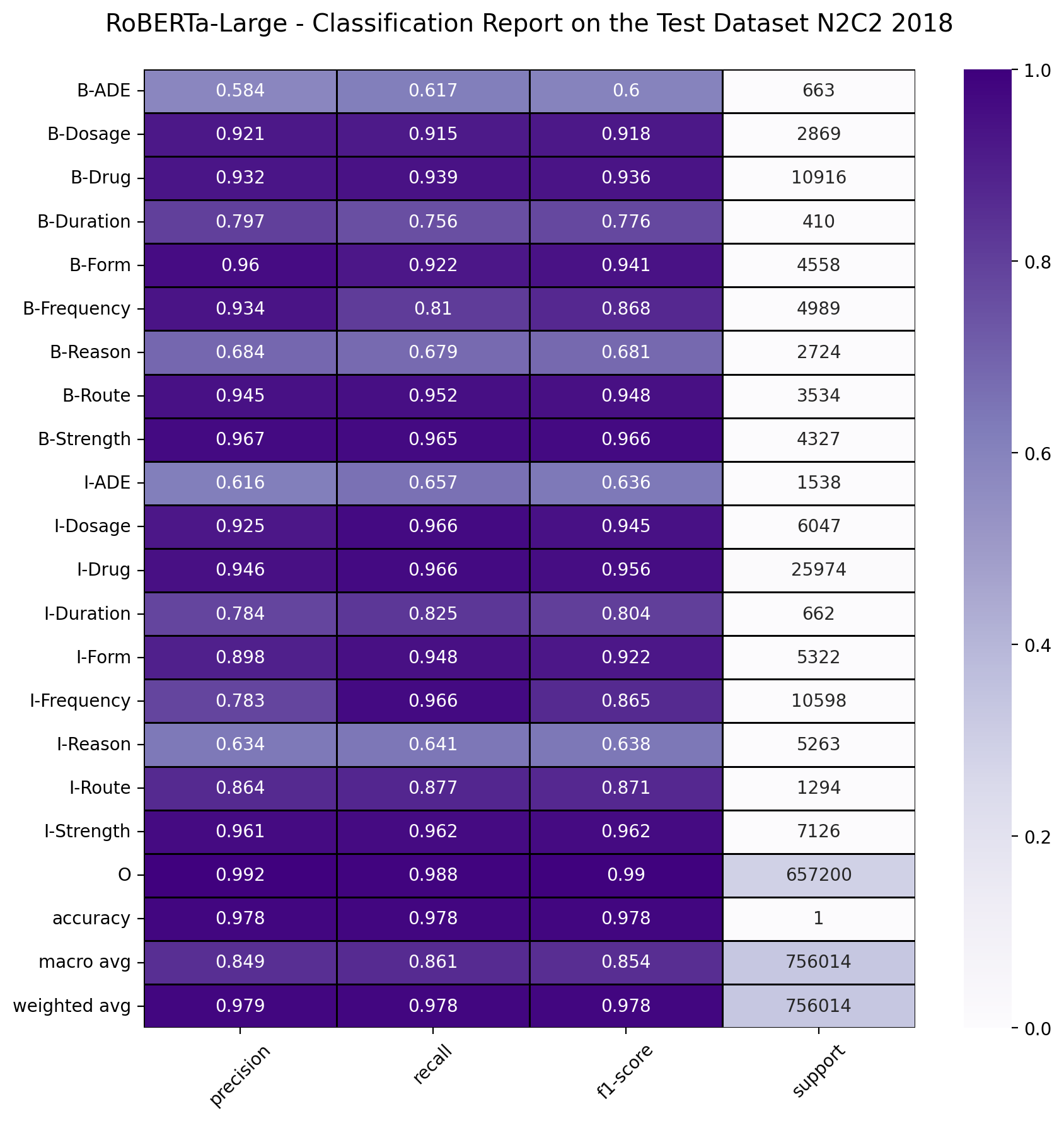}
    \caption{RoBERTa-L Eval at Sub-word Level on n2c2 2018 test data.}
    \label{fig:RoBERTa-Large_token_bios}
\end{figure*}

\begin{figure*}[]
\centering
\includegraphics[width=0.99\textwidth]{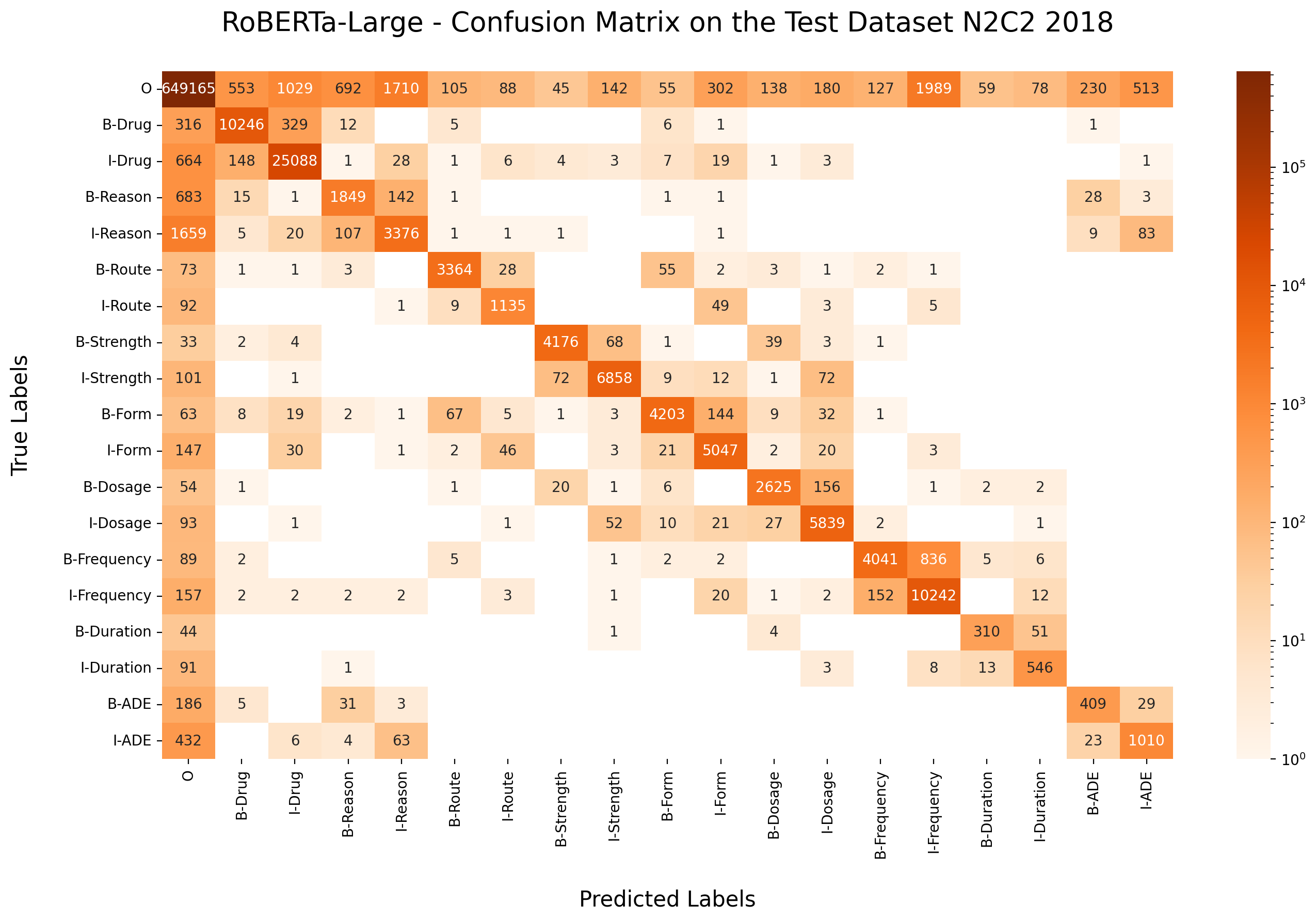}
    \caption{RoBERTa-L Eval Confusion Matrix at Sub-word Level on n2c2 2018 test data.}
    \label{fig:RoBERTa-Large_token_confusion_matrix}
\end{figure*}

\begin{figure*}[]
\centering
\includegraphics[width=0.99\textwidth]{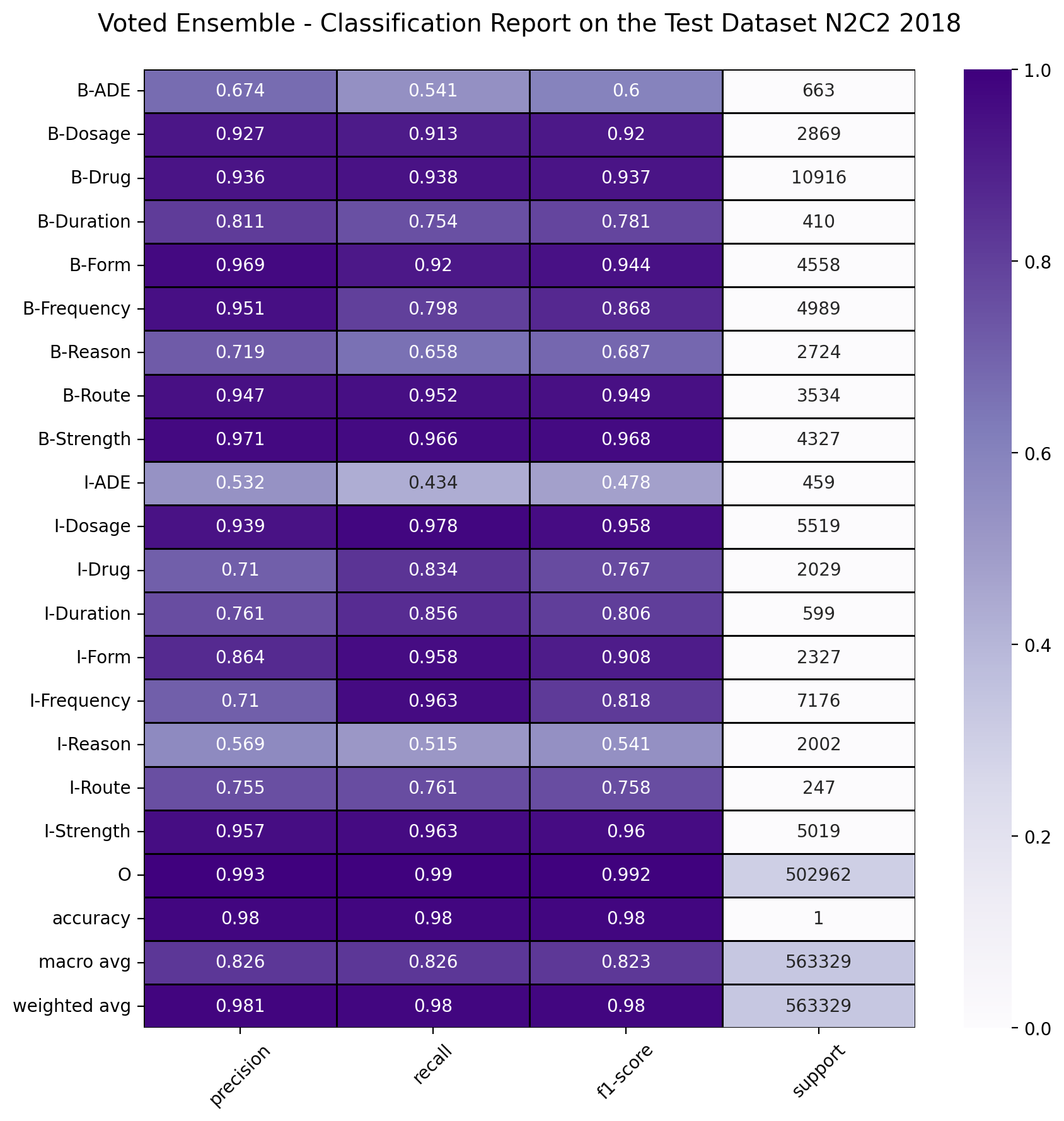}
    \caption{word-level grouping ensemble, max logit voting Eval on n2c2 2018 test data.}
    \label{fig:max-Logit_Ensemble_Voting_word_bios}
\end{figure*}

\begin{figure*}[]
\centering
\includegraphics[width=0.99\textwidth]{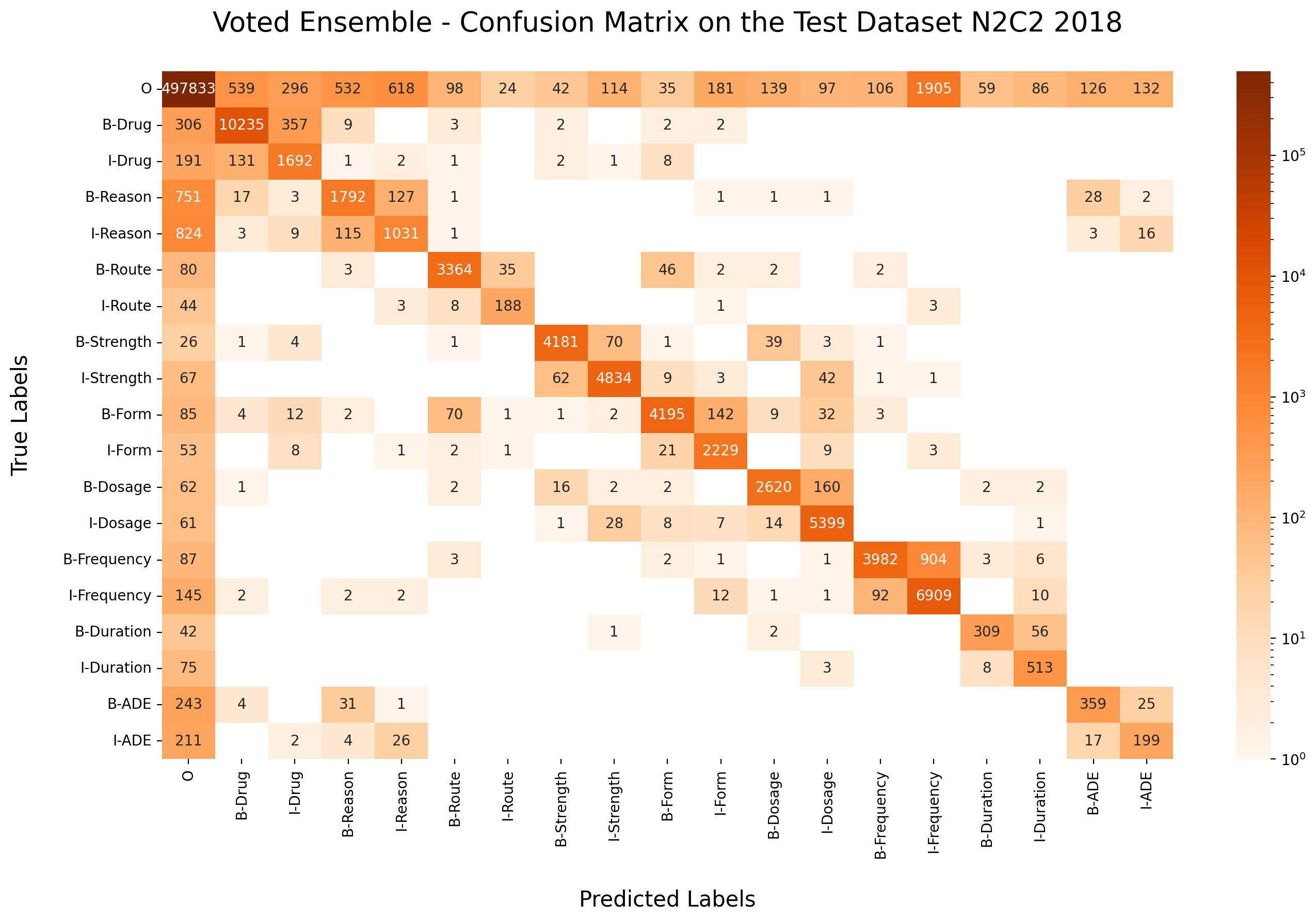}
    \caption{word-level ensemble max-logit votingEval confusion matrix on n2c2 2018 test data.}
    \label{fig:max-Logit_Ensemble_Voting_word_confusion_matrix}
\end{figure*}

\begin{figure*}[]
\centering
\includegraphics[width=0.99\textwidth]{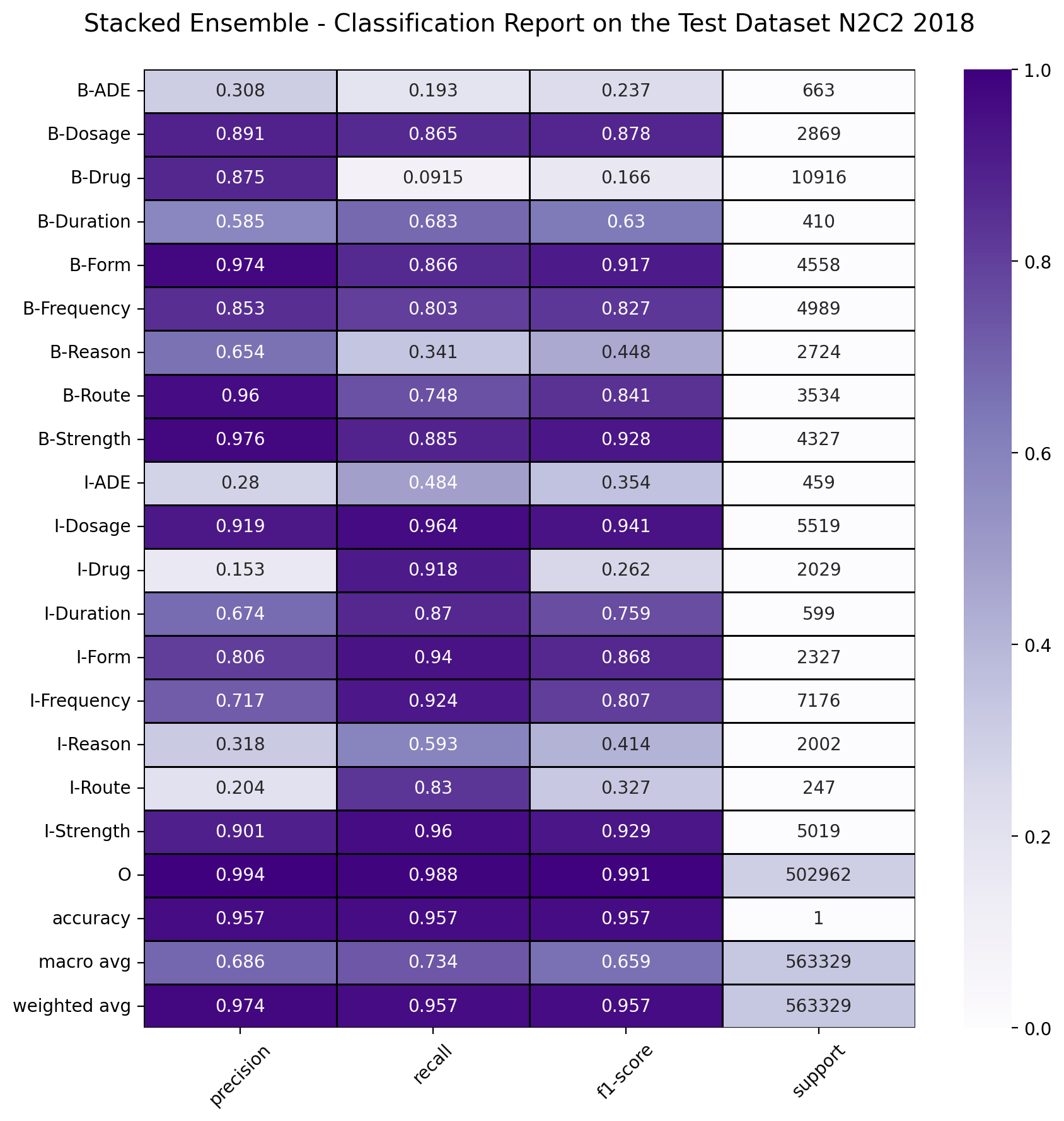}
    \caption{word-level grouping ensemble, max logit (logits, non-one-hot): stacked ensemble Eval on n2c2 2018 test data, which is much lower than the max voting.}
    \label{fig:max_Stacked_Ensemble_word_bios}
\end{figure*}

\begin{figure*}[]
\centering
\includegraphics[width=0.99\textwidth]{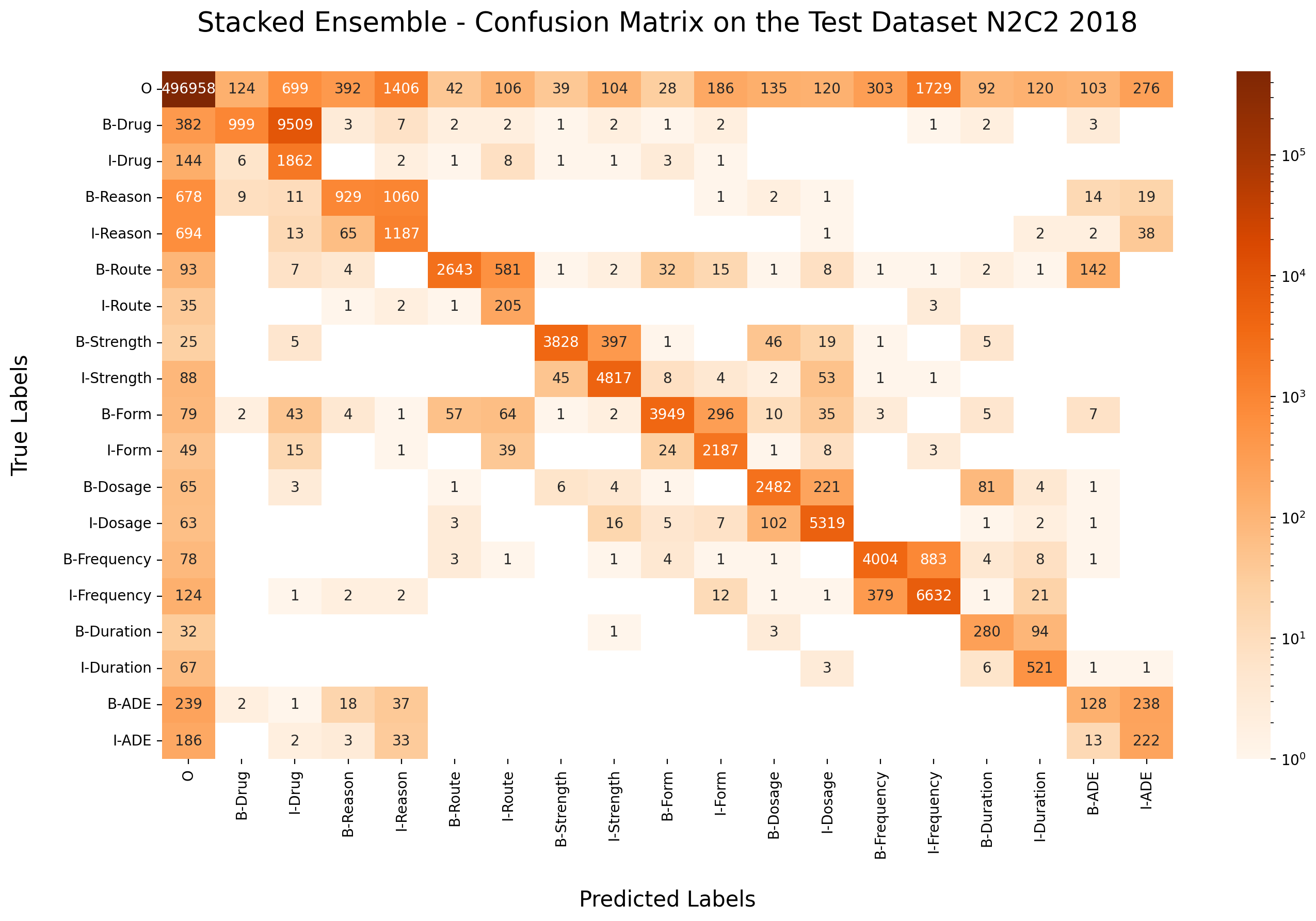}
    \caption{word-level grouping ensemble, max logit: stacked ensemble confusion matrix Eval on n2c2 2018 test data, which is much worse than the max voting.}
    \label{fig:max_Stacked_Ensemble_word_confusion_matrix}
\end{figure*}

\subsection{Individual vs Ensemble Models}
The word-level performance comparisons from individual models and voting max-logit ensembles are presented in Table \ref{tab:individual-vs-ensemble-max-voting-word-level}.

\section{Huggingface Models used}
\textsc{InsightBuddy} integrated individual models and their Huggingface repositories are listed in Table \ref{tab:EnsembleNER-model-list}.

\section{Hyper Parameter Optimisations}
\label{sec:appendix_param}

We used a set of parameters for model fine-tuning and selected the better parameter set as below using the validation data. We tried different learning rates (0.0001, 0.0002, 0.00005) and batch sizes (16, 32).

\begin{itemize}
    \item learning\_rate: 0.00005
    \item train\_batch\_size: 32
    \item eval\_batch\_size: 32
    \item seed: 42
    \item optimizer: Adam with betas=(0.9,0.999) and epsilon=1e-08
    \item lr\_scheduler\_type: linear
    \item lr\_scheduler\_warmup\_ratio: 0.1
    \item num\_epochs: 4
    \item mixed\_precision\_training: Native AMP
\end{itemize}





\onecolumn
\newpage


\clearpage

\pagebreak
\begin{longtable}{|l|c|c|c|}
\hline
\multicolumn{4}{|c|}{\textbf{Individual models max-logit grouping (word)}} \\
\hline
\textbf{Metric}          & \textbf{P}    & \textbf{R}    & \textbf{F1}   \\
\hline
\endfirsthead
\hline
\textbf{Model}                   & \textbf{P}    & \textbf{R}    & \textbf{F1}   \\
\hline
\endhead
\hline
\endfoot

\multicolumn{4}{|c|}{\textbf{BERT }} \\
\hline
accuracy                 & \multicolumn{3}{c|}{0.9773} \\\hline 
macro avg                & 0.7942 & 0.7965 & 0.7928 \\
weighted avg             & 0.9784 & 0.9773 & 0.9775 \\
\hline

\multicolumn{4}{|c|}{\textbf{RoBERTa }} \\
\hline
accuracy                 & \multicolumn{3}{c|}{0.9780} \\\hline 
macro avg                & 0.8029 & 0.8201 & 0.8094 \\
weighted avg             & 0.9795 & 0.9780 & 0.9784 \\
\hline

\multicolumn{4}{|c|}{\textbf{RoBERTa-Large }} \\
\hline
accuracy                 & \multicolumn{3}{c|}{0.9788} \\\hline 
macro avg                & 0.8091 & 0.8351 & 0.8202 \\
weighted avg             & 0.9802 & 0.9788 & 0.9792 \\
\hline

\multicolumn{4}{|c|}{\textbf{ClinicalBERT }} \\
\hline
accuracy                 & \multicolumn{3}{c|}{0.9780} \\\hline 
macro avg                & 0.8087 & 0.7916 & 0.7964 \\
weighted avg             & 0.9785 & 0.9780 & 0.9779 \\
\hline

\multicolumn{4}{|c|}{\textbf{BioBERT }} \\
\hline
accuracy                 & \multicolumn{3}{c|}{0.9776} \\\hline 
macro avg                & 0.7972 & 0.8131 & 0.8027 \\
weighted avg             & 0.9787 & 0.9776 & 0.9779 \\
\hline

\multicolumn{4}{|c|}{\textbf{BioClinicalBERT }} \\
\hline
accuracy                 & \multicolumn{3}{c|}{0.9776} \\\hline 
macro avg                & 0.7999 & 0.8090 & 0.8017 \\
weighted avg             & 0.9788 & 0.9776 & 0.9779 \\
\hline

\multicolumn{4}{|c|}{\textbf{BioMedRoBERTa }} \\
\hline
accuracy                 & \multicolumn{3}{c|}{0.9783} \\\hline 
macro avg                & 0.8065 & 0.8224 & 0.8122 \\
weighted avg             & 0.9797 & 0.9783 & 0.9786 \\
\hline

\multicolumn{4}{|c|}{\textbf{PubMedBERT }} \\
\hline
accuracy                 & \multicolumn{3}{c|}{0.9784} \\\hline 
macro avg                & 0.8087 & 0.8292 & 0.8166 \\
weighted avg             & 0.9800 & 0.9784 & 0.9788 \\
\hline

\multicolumn{4}{|c|}{\textbf{Voting Max logit ensemble word level}} \\
\hline
accuracy                 & \multicolumn{3}{c|}{0.9796} \\\hline 
macro avg                & \textbf{0.8261} & 0.8259 & \textbf{0.8232} \\
weighted avg             & 0.9807 & 0.9796 & 0.9798 \\
\hline\caption{word-level individual model (grouping using max-logit) vs ensemble using max-logit, Eval on n2c2 2018 test data}
    \label{tab:individual-vs-ensemble-max-voting-word-level}
\end{longtable}

\begin{table}[t]
\centering
\begin{tabular}{|l|c|c|c|c|c|}
\hline
Model & Macro P & Macro R & Macro F & Accuracy & Tokens(sub-words) \\
\hline
BERT & 0.8336 & 0.8264 &0.8283 & 0.9748 & 756798 \\
ROBERTa & 0.8423 & 0.8471 & 0.8434 & 0.9770 & 756014 \\
ROBERTa-L & \textbf{0.8489} & \textbf{0.8606} & \textbf{0.9782} & \textbf{0.8538} & 756014 \\\hline 
PubMedBERT & 0.8324 & 0.8381 &0.8339 & \textbf{0.9783} & 681211 \\
ClinicalBERT & \textbf{0.8482} & 0.8245 & 0.8341 & 0.9753 & \textit{796313} \\
BioMedRoBERTa & \textbf{0.8482} & \textbf{0.8477} & \textbf{0.8468} & 0.9775 & 756014 \\
BioClinicalBERT & 0.8440 & 0.8405 & 0.8406 & 0.9751 & 791743 \\
BioBERT & 0.8365 & 0.8444 &0.8393 & 0.9750 & 791743 \\
\hline
\end{tabular}
    \caption{\textsc{InsightBuddy} individual sub-word level model eval on n2c2-2018 test set. The first group: normal domain PLM; The second group: biomedical PLM. The different numbers of Support are due to the different tokenizers they used -- ROBERTa and ROBERTa-L use the same tokenizers, BioClinicalBERT and BioBERT use the same tokenizers, and other models all use different tokenizers; PubMedBERT generated the least number of sub-words/tokens 681,211 while ClinicalBERT generated the largest number of tokens 796,313.}
    \label{tab:individual-sub-word-level-models}
\end{table}

\begin{longtable}{|l|l|}
\hline
\textbf{Ensemble List} & \textbf{Link} \\ \hline
BERT & \href{https://huggingface.co/google-bert/bert-base-uncased}{https://huggingface.co/google-bert/bert-base-uncased} \\ \hline
BioBERT & \href{https://huggingface.co/dmis-lab/biobert-base-cased-v1.2}{https://huggingface.co/dmis-lab/biobert-base-cased-v1.2} \\ \hline
ClinicalBERT & \href{https://huggingface.co/medicalai/ClinicalBERT}{https://huggingface.co/medicalai/ClinicalBERT} \\ \hline
BioClinicalBERT & \href{https://huggingface.co/emilyalsentzer/Bio\_ClinicalBERT}{https://huggingface.co/emilyalsentzer/Bio\_ClinicalBERT} \\ \hline
PubMedBERT & \href{https://huggingface.co/microsoft/BiomedNLP-BiomedBERT-base-uncased-abstract-fulltext}{https://huggingface.co/microsoft/BiomedNLP-BiomedBERT-base-uncased-abstract-fulltext} \\ \hline
BioMedRoBERTa & \href{https://huggingface.co/allenai/biomed_roberta_base}{https://huggingface.co/allenai/biomed\_roberta\_base} \\ \hline
RoBERTa & \href{https://huggingface.co/FacebookAI/roberta-base}{https://huggingface.co/FacebookAI/roberta-base} \\ \hline
RoBERTa Large & \href{https://huggingface.co/FacebookAI/roberta-large}{https://huggingface.co/FacebookAI/roberta-large} \\ \hline
    \caption{\textsc{InsightBuddy} integrated individual models and their Huggingface repositories. }
    \label{tab:EnsembleNER-model-list}
\end{longtable}

\end{document}